\pdfoutput=1

\documentclass[11pt]{article}

\usepackage{ACL2023}

\usepackage{times}
\usepackage{latexsym}
\usepackage{booktabs}
\usepackage{amsmath}
\usepackage{enumitem}
\usepackage{graphicx,ragged2e}
\usepackage{caption}

\usepackage[T1]{fontenc}

\usepackage[utf8]{inputenc}

\usepackage{microtype}

\usepackage{inconsolata}

\usepackage{graphicx}
\usepackage{subcaption}
\title{American Sign Language Handshapes Reflect \\ Pressures for Communicative Efficiency} 

\author{Kayo Yin \\
  UC Berkeley \\ 
  \\ \And
  Terry Regier \\
  UC Berkeley \\
  \texttt{\{kayoyin, terry.regier, klein\}@berkeley.edu}\\ \And
  Dan Klein \\
  UC Berkeley \\
  \\}

\begin{document}
\maketitle

\begin{abstract}


Communicative efficiency is a key topic in linguistics and cognitive psychology, with many studies demonstrating how the pressure to communicate with minimal effort guides the form of natural language. However, this phenomenon is rarely explored in signed languages. This paper shows how handshapes in American Sign Language (ASL) reflect these efficiency pressures and provides new evidence of communicative efficiency in the visual-gestural modality.

We focus on hand configurations in native ASL signs and signs borrowed from English to compare efficiency pressures from both ASL and English usage. First, we develop new methodologies to quantify the articulatory effort needed to produce handshapes and the perceptual effort required to recognize them. Then, we analyze correlations between communicative effort and usage statistics in ASL or English. Our findings reveal that frequent ASL handshapes are easier to produce and that pressures for communicative efficiency mostly come from ASL usage, rather than from English lexical borrowing.\footnote{Data and code at \url{https://github.com/kayoyin/asl-efficiency}}

\end{abstract}

\section{Introduction}\label{sec:intro}

There is increasing evidence suggesting that human languages adapt to the needs of their users by minimizing the effort required by the sender and receiver to achieve successful communication.  It is argued that natural languages tend to find an optimal balance between the efforts of the two participants when they are at odds with each other \cite{zipf2016human, king1967functional, piantadosi2011word, gibson2019efficiency, rasenberg2022multimodal}. 

\begin{figure}[ht]
\centering
    \includegraphics[width=\linewidth]{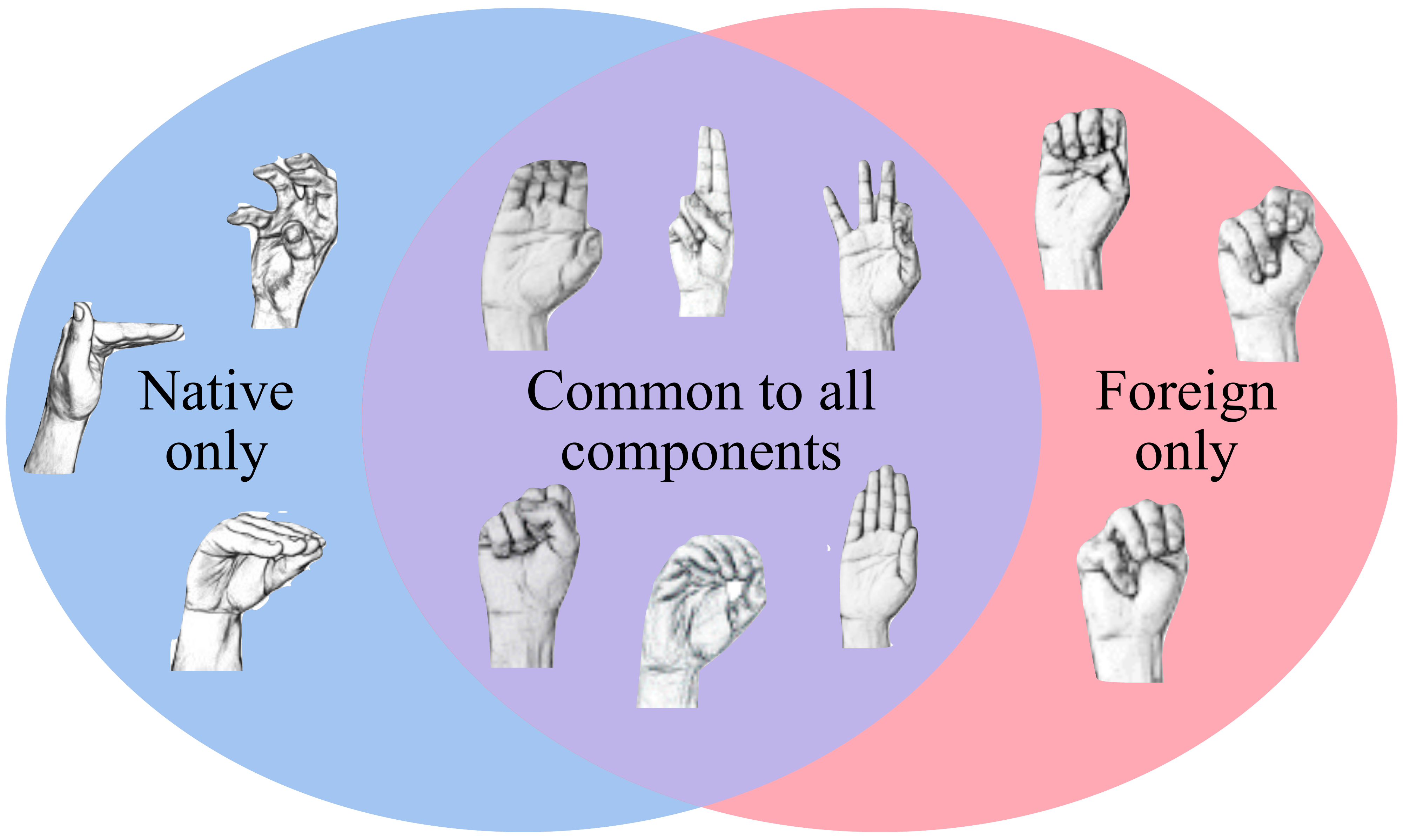}
  \caption{Examples of handshapes in ASL components. The ASL lexicon can be divided into a native component (e.g. signs native to ASL; left) and a foreign component (e.g. fingerspelling, loan signs; right). 19 out of 22 handshapes in ASL fingerspelling also appear in the native lexicon\protect\footnotemark.}
  \label{fig:fig1}
\end{figure}

While the majority of existing work studying communicative efficiency has focused on spoken languages, recent results have provided evidence that signed languages are also shaped by drives for efficient communication in the visual-gestural modality. For example,  \citet{napoli2011some} found that casual signing prioritizes moving fewer, more distal joints, increasing articulatory ease. \citet{caselli2022perceptual} found that rare hand configurations are produced close to the signer's face, increasing perceptual ease.\footnotetext{Handshape artworks shared with permission from \citet{vicars}.}

This paper further explores communicative pressures on visual signed languages by focusing on American Sign Language (ASL) handshapes. Handshapes refer to distinctive configurations of the hand and fingers, and are one of the five fundamental parameters characterizing signs in ASL (\S\ref{sec:phonology}). A finite set of handshapes is combined with different movements, locations, palm orientations, and non-manual markers to express various ASL signs. Figure \ref{fig:fig1} shows example handshapes, some of which are used only in native ASL signs (left), some of which are used only in ASL signs borrowed from other languages (right), and some of which are used in both contexts (center).

We investigated evidence of communicative efficiency in both the \textit{native} and \textit{foreign} components of ASL. \textit{Foreign} signs are borrowed from other languages in contrast to \textit{native} signs that are inherently derived from ASL itself.\footnote{Although signs originating from lexical borrowing are technically a part of ASL, we refer to them as the ``foreign'' component for our purposes, following \citet{brentari2001language}. The foreign component of ASL includes signs borrowed from English as well as signs borrowed from other signed languages (e.g. country signs such as JAPAN, CHINA). In this paper, we only address foreign signs that derive from English.} 
The foreign component of ASL includes \textit{neutral fingerspelling} where English words are spelled out using one-handed signs that each represent a letter of the English alphabet (\S\ref{sec:fingerspelling}); \textit{loan signs} where commonly fingerspelled words evolved into a lexicalized sign; and \textit{initialized signs} that are produced using the handshape of the first letter of its English translation.


\begin{figure}[ht]
\centering
    \includegraphics[width=\linewidth]{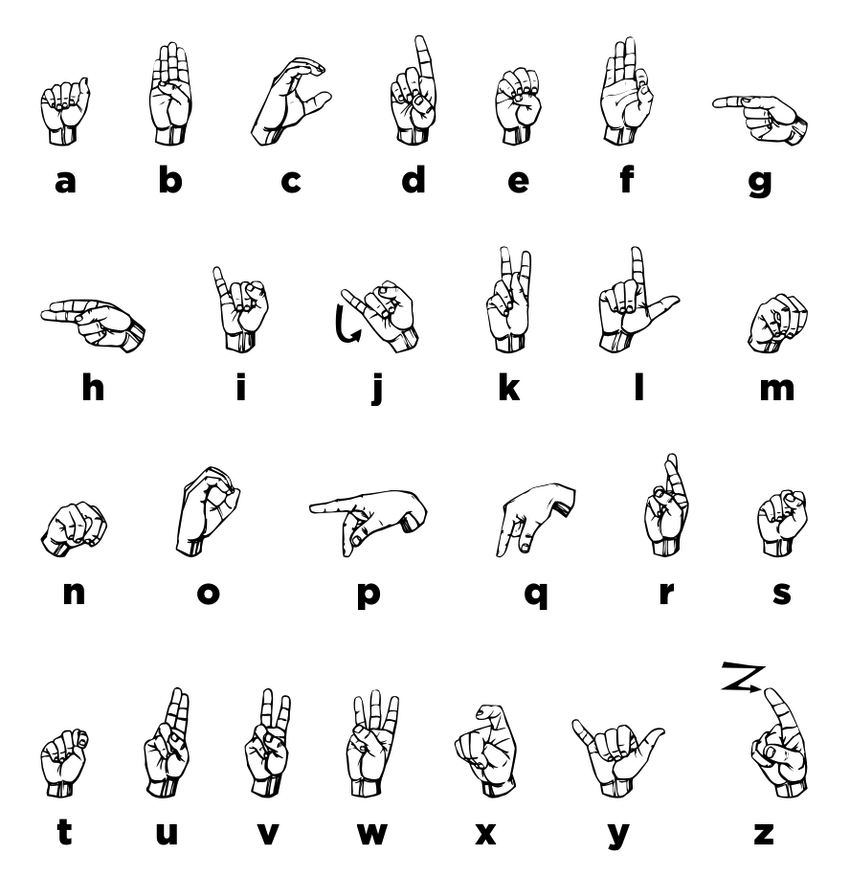}
  \caption{ASL fingerspelling letters from \citet{keane2014towards}. For certain letters, the palm is oriented differently than in practice to better illustrate the handshape.}
  \label{fig:alphabet}
\end{figure}


To compare effects of ASL and English usage on handshape efficiency, we focused on fingerspelling handshapes (\textit{FS handshapes}; Figure \ref{fig:alphabet}) since 19 out of 22 FS handshapes appear in both the native and foreign components. We were motivated by the following research questions:

\begin{enumerate}
    \item[\textbf{RQ1}]Do handshapes reflect pressure for communicative efficiency?
    \item[\textbf{RQ2}]Is pressure for efficiency mostly or all from ASL usage, or does English usage also play a role?
\end{enumerate}

To test these ideas, we designed new methodologies to measure \textit{articulatory effort} required by the sender to produce handshapes, as well as the \textit{perceptual effort} needed by the receiver to recognize handshapes. We propose three predictions following from the hypothesis of communicative efficiency (\textbf{P1-3}). 

First, we predicted that according to the hypothesis of communicative efficiency, FS handshapes which appear frequently in native ASL signs should be easier to produce: this would help to keep overall articulatory (sender) effort low (\textbf{P1}). Our evidence indeed supports this prediction: we found a positive correlation between handshape frequency in native ASL signs and articulatory ease.

Next, since foreign signs obey fewer phonological constraints observed in ASL \cite{brentari2001language}, one might expect that handshapes in foreign signs reflect little to no pressure for communicative efficiency. If so, letters that appear frequently in English do not necessarily correspond to being easier to sign in ASL fingerspelling (\textbf{P2}). We indeed found no significant correlation between English letter frequency and fingerspelling articulatory ease, which supports \textbf{P2}.

If English usage has negligible effect on communicative efficiency, foreign signs should reflect little to no pressure for efficiency in \textbf{perceptual (receiver) effort} either (\textbf{P3}). If there are perceptual pressures for efficiency, pairs of letters that appear in similar contexts in English should be easier to disambiguate from one another in ASL fingerspelling: this would help to keep receiver effort low by placing perceptually disambiguating elements in places that might otherwise be confusable. We indeed verified that there is no correlation between English letter confusability and perceptual ease in ASL fingerspelling, which supports \textbf{P3}.

In summary, our analysis finds evidence of pressure for articulatory ease in ASL handshapes and suggests that pressure for communicative efficiency mostly comes from ASL usage, not from English usage.

\section{Background}

\subsection{Structure of ASL}\label{sec:phonology}
ASL signs are often subdivided into five parameters: handshape, palm orientation, movement, location of articulation, and non-manual markers \cite{stokoe, battison1978lexical, liddell1989american}. Sign morphemes, unlike spoken morphemes, are usually simultaneously produced instead of sequentially. 
It is estimated that there are about 45 different handshapes \cite{battison1978lexical}, however, some of these handshapes are phonemically identical and interchangeable \cite{tennant1998american}. 

\citet{brentari2001language} proposed that the ASL lexicon is stratified into a core lexicon that is \textit{native} to ASL and a \textit{foreign} component with varying distances from the core lexicon. The foreign component includes vocabulary such as \textit{fingerspelling} where English words are spelled letter by letter using one-handed signs (Figure \ref{fig:alphabet}); \textit{loan signs} where commonly fingerspelled words evolved into a lexicalized sign; and \textit{initialized signs} that are produced using the handshape of the first letter of its English translation (e.g. CLASS and FAMILY are signed with the handshapes for C and F respectively). As forms diverge further away from the core component, they obey fewer of the phonological constraints observed in the core component: for example, ASL phonotactic rules \cite{perlmutter1993sonority, brentari1998prosodic, battison1978lexical} hold for core ASL vocabulary but are almost entirely violated in fingerspelled words.

Handshapes in ASL vary in \textit{markedness}, where unmarked handshapes are defined as handshapes that are easier to learn, produce, and process. To determine the hierarchy of markedness in handshapes, linguists have examined various features such as motoric complexity \cite{braem1990acquisition, ann2006frequency}, child language acquisition \cite{siedlecki1997young}, and visual perception \cite{lane1976preliminaries}. Although there is no consensus on the exact markedness hierarchy, unmarked handshapes are usually limited to 7 handshapes (B, A, S, C, O, 1, 5; \cite{battison1978lexical}). 

Our proposed measure for articulatory effort is based on notions of motoric complexity. However, instead of dividing handshapes into (around 4) categories of markedness, we measured articulatory effort on a continuum, and we disentangled articulatory effort from other features included in markedness such as ease of learning.

\subsection{Fingerspelling}\label{sec:fingerspelling}

Fingerspelling results from language contact between a signed language and a surrounding spoken language written form \cite{battison1978lexical,wilcox1992phonetics,brentari2001language, patrie2011fingerspelled}. In fingerspelling, a set of manual gestures corresponds to a written orthography or phonetic system of a spoken language and can be comprehended visually or tactilely. Fingerspelling is often used to indicate names and technical concepts for which no conventional signs exist. It is also sometimes used when there are equivalent signs in the signed language, for example for linguistic strategies such as emphasis or style \cite{padden1998asl,padden2006learning, montemurro2018emphatic}. 


In ASL fingerspelling, each letter of the English alphabet is represented by one of 22 unique handshapes, plus, for certain letters, a non-default palm orientation (G, H, P, Q) or an added movement (J,Z; Figure \ref{fig:alphabet}). In our analysis, we focused on communicative efficiency in handshapes only to allow comparison between linguistic forms in ASL fingerspelling and lexical signs: only two non-default handshapes and two movements appear in fingerspelling, which makes it difficult to extrapolate analysis to include movement and palm orientation.



\section{Data}\label{sec:data}
We describe the data we use to compute usage statistics in ASL and English, as well as to compute the articulatory and perceptual effort in ASL handshapes.

\subsection{ASL-LEX}
We used ASL-LEX \cite{caselli2017asl, sehyr2021asl} to extract usage statistics of ASL signs (for \textbf{P1}). ASL-LEX is a public database of 2,723 ASL signs annotated with their lexical and phonological properties. Notably, ASL-LEX is annotated with the handshape category of each sign, as well as sign frequency ratings by 25 deaf ASL signers, and whether the sign is an initialized sign, a fingerspelled loan sign, or a native sign. 

\subsection{English}\label{sec:english}
To calculate usage statistics in English (for \textbf{P2} and \textbf{P3}), we used a selection of Wikipedia articles in English. We randomly sampled 10,000 articles from the Wikipedia dataset on Hugging Face \cite{lhoest2021datasets}, then we sorted all words in the selected articles (case normalized) by frequency. To better approximate words that are likely to be fingerspelled, we discarded the 20,000 most frequent types (since common English words are unlikely to be fingerspelled) and all words that appear only once to reduce noise. We used the remaining 71,785 words to compute usage metrics in \S\ref{sec:usage}. 

\begin{figure}
\centering
    \includegraphics[width=\linewidth]{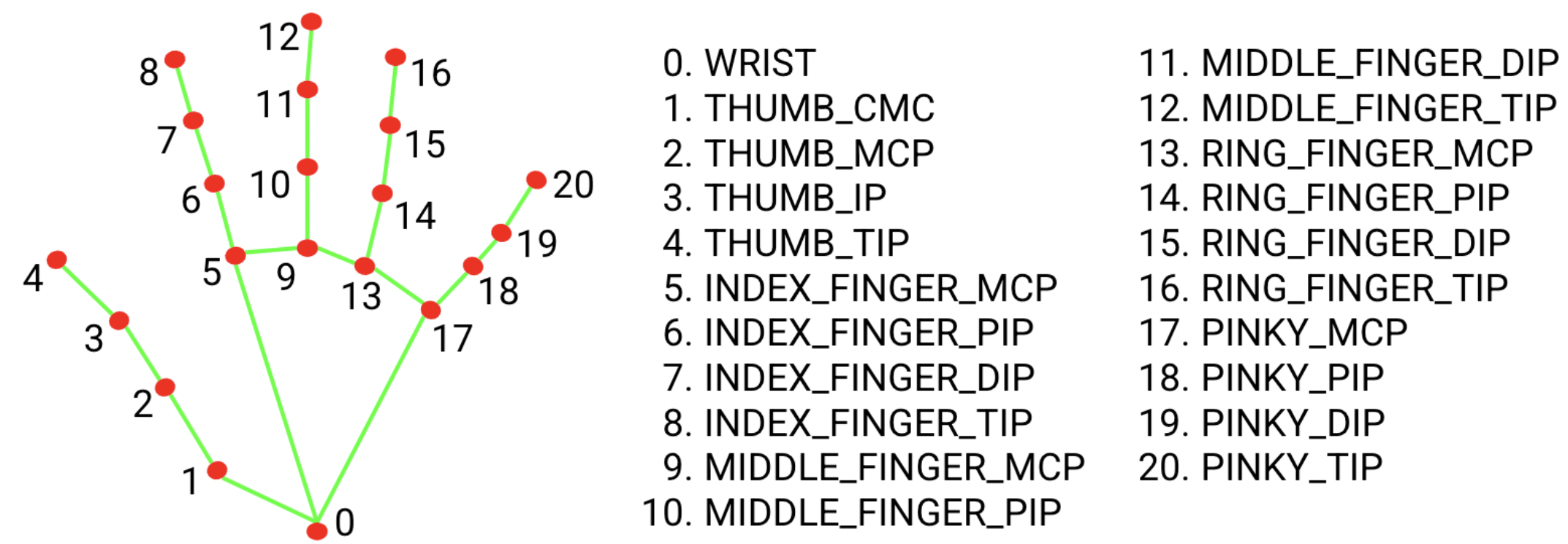}
  \caption{List of hand landmarks in MediaPipe \cite{zhang2020mediapipe}.}
  \label{fig:landmarks}
\end{figure}

\subsection{ASL Fingerspelling}

We use FS handshapes to compare pressures for communicative efficiency in native and foreign components of ASL. To our knowledge, there were no existing public datasets of individual letters from ASL fingerspelling recorded from native or fluent ASL signers.
We, therefore, collected data by extracting fingerspelled letters from the ASL Fingerspelling Recognition Corpus \cite{asl-fingerspelling}. This corpus consists of hand, body, and face landmarks extracted using MediaPipe \cite{lugaresi2019mediapipe} from $116,170$ videos of English phrases, addresses, phone numbers, and URLs, fingerspelled in ASL by over $100$ Deaf ASL signers. 

However, this dataset only contains English phrases and the fingerspelling video for the phrase, it does not contain character-level annotation for each fingerspelled letter. We therefore extracted static frames of fingerspelled letters using a heuristic algorithm and manual post-correction: given an English phrase $P$ with $n$ letters and a sequence of hand landmarks $H$ corresponding to the fingerspelling of $P$, we first computed the velocity of transitions between frames by taking the difference between positional coordinates of each hand landmark in consecutive frames. Then, we extracted $n$ frames from $H$ that have the sharpest local minima in transition velocity. Intuitively, the hand slows down between two letter transitions to produce the letter. We then aligned the $n$ letters in $P$ with the $n$ extracted frames in order, and manually corrected errors in alignment.

\begin{table}[htp]
    \centering
      \resizebox{\linewidth}{!}{
    \begin{tabular}{ccc|ccc}
    \toprule
     Letter & Count & \% & Letter & Count & \%\\
     \midrule
A & 90 & 8.5 & N & 74 & 7.0\\
B & 21 & 2.0 & O & 72 & 6.8\\
C & 25 & 2.4 & P & 20 & 1.9\\
D & 49 & 4.6 & Q & 10 & 0.9\\
E & 100 & 9.4 & R & 65 & 6.1\\
F & 25 & 2.4 & S & 53 & 5.0\\
G & 27 & 2.5 & T & 80 & 7.5\\
H & 60 & 5.6 & U & 29 & 2.7\\
I & 82 & 7.7 & V & 20 & 1.9\\
K & 15 & 1.4 & W & 37 & 3.5\\
L & 51 & 4.8 & X & 13 & 1.2\\
M & 21 & 2.0 & Y & 23 & 2.2\\
     \bottomrule
    \end{tabular}}
    \caption{Distribution of letters extracted from the ASL Fingerspelling Recognition Corpus.}
    \label{tab:fs-data}
\end{table}

We collected a total of 1,062 static, isolated letters\footnote{We exclude J and Z from our study since these two letters require movement. J shares the same handshape as I and Z shares the same handshape as D.} (Table \ref{tab:fs-data}).
Unlike ASL signs, ASL fingerspelling is expressed with the dominant hand only, therefore we discarded landmarks from the body, face, and non-dominant hand. For each handshape, we kept the 21 landmarks from the dominant hand (Figure \ref{fig:landmarks}).




\section{Methodology}\label{sec:method}
We now describe our methodology to measure usage metrics in ASL and English, and articulatory/perceptual effort in handshapes.

\subsection{Usage Metrics}\label{sec:usage}
\subsubsection*{Frequency}
For handshape frequency in ASL, we summed the frequency ratings of signs in ASL-LEX that have the target FS-handshape as the handshape of the dominant hand in the first morpheme. ASL-LEX contains a total of 1,204 signs with FS handshapes, among which 903 are native signs, 271 are initialized signs, and 30 are loan signs.

For letter frequency in our English corpus, we counted the occurrence of each letter over rare English words in \S\ref{sec:english}, multiplied by the frequency of the word.

\subsubsection*{Confusability by context}
We quantified how \textit{confusable by context} two letters are, that is, how often two letters appear in similar contexts in English, by measuring the conditional entropy $H(X|C)$ for two letters $X=\{x_1, x_2\}$ given context $C$. We define the context of a letter as the character $n$-grams preceding the letter within a word: the context of the letter at position $i$ of a word $w$ is $c_i = w[\max(0, i-4):i]$. Contexts are of length between 1-4 letters. 

We approximated $p(x_i, c_i)$ and $p(c_i)$ by using an $n$-gram character model over the word list from \S\ref{sec:english}. Pairs of letters $x_1, x_2$ with \textbf{high conditional entropy} $H(X|C)$ are not well disambiguated by context $C$, therefore, \textbf{more confusable by context}.

\begin{figure}
\centering
    \includegraphics[width=\linewidth]{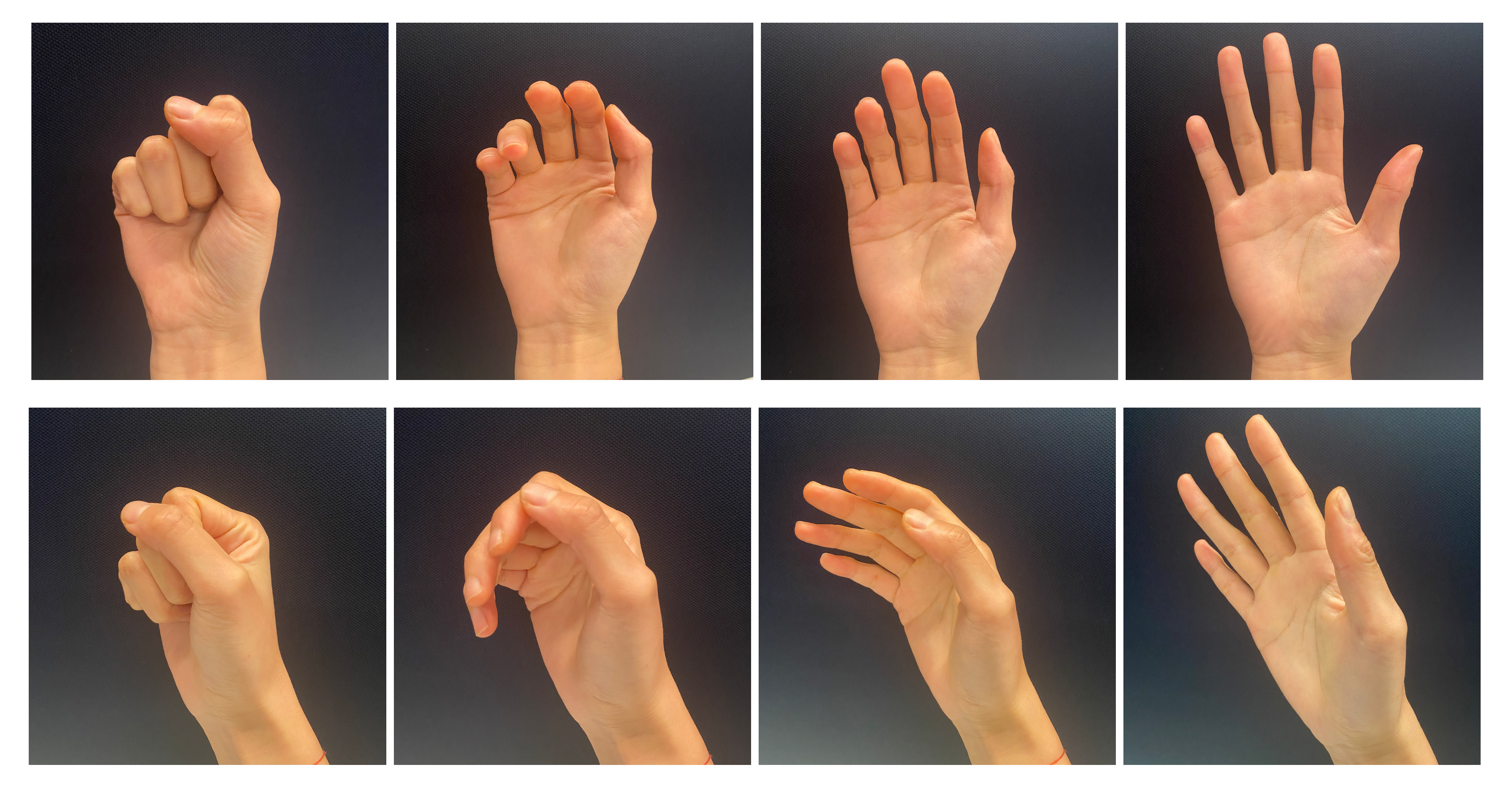}
  \caption{Hands in low effort positions.}
  \label{fig:resthands}
\end{figure}

\subsection{Measures of Effort}

For the following measures of effort, we computed the distance between joint angles, which we define as follows:
\(
D(\alpha, \beta) = |\alpha-\beta| \text{ mod } 2\pi
\)
where $\alpha$ and $\beta$ are two joint angles.


\subsubsection*{Joint angle representation}\label{sec:rot}
To obtain joint angles, we first applied a change of basis for the hand landmarks of our ASL fingerspelling data from 3D cartesian coordinates to the joint rotational space. 
For each set of three adjacent hand landmarks, we calculated the 3D angle between the two vectors defined by the three points. We thus obtain an angular representation of each handshape consisting of the set of angles of each joint in the hand. This rotational representation is invariant to translation, reflection, and scaling, which is desired for analyzing handshapes, and removes the need to normalize our data.

\subsubsection*{Finger independence}
We approximated the effort required to produce a certain handshape by measuring finger independence, where the higher the finger independence, the more different finger positions are from each other, and thereby more effort is often required to produce the handshape. For example, A and C in Figure \ref{fig:alphabet} have low finger independence, P and V have high finger independence. This metric is based on studies in kinematics and neurophysiology that have demonstrated high finger independence requires more articulatory effort \cite{hager2000quantifying, schieber2004hand, lang2004human, santello2016hand}. 

We defined groups of the same type of joints (metacarpophalangeal, proximal interphalangeal, or distal interphalangeal) across fingers as
\begin{gather*}
\mathcal{G}_\mathcal{J} = \{\{5,9,13,17\},\{6,10,14,19\}, \\
\{7,11,15,19\}\}
\end{gather*}
where the joint indices follow MediaPipe landmarks in Figure \ref{fig:landmarks}. We excluded thumb joints from $\mathcal{G}_\mathcal{J}$ because the thumb has the highest digit individuation, and the position of the thumb does not allow a direct comparison between its joint angles with the other finger joint angles over MediaPipe landmarks.

We took into account articulatory effort from the thumb by defining \textit{thumb effort}. To do so, we first collected images of hands at four different low-effort positions from frontal and side views, which we hereby call `resting hands' (Figure \ref{fig:resthands}). We then extracted hand landmarks using MediaPipe and calculate joint angles in resting hands. Then, thumb effort is defined as:
$$
TE(\text{hand}) = \min_{\text{hand}_\text{r} \in \mathcal{R}} \sum_{\alpha \in \mathcal{T}_\text{hand}, \beta \in \mathcal{T}_{\text{hand}_\text{r}}} D(\alpha, \beta) / N
$$
where we took the minimum over the set of resting hands $\mathcal{R}$: for each resting hand, we took the mean over the distances between each joint in the thumb of the resting hand $\mathcal{T}_{\text{hand}_\text{r}}$ and the thumb of the hand we are measuring $\mathcal{T}_\text{hand}$. The more different the thumb configuration is from thumbs in resting hands, the more effort is required to place the thumb in such configuration.

We then define finger independence as follows:
\begin{align*}
FI(\text{hand}) &=  2 TE(\text{hand}) \quad \quad\quad\quad \\
&+ \sum_{\mathcal{J} \in \mathcal{G}_\mathcal{J}} \sum_{i,j \in \mathcal{J} | i < j} D(\alpha_i, \alpha_j)
\end{align*}

where for each group of joints $\mathcal{J} \in \mathcal{G}_\mathcal{J}$ in a hand, we computed the pairwise distance between each joint angle in $\mathcal{J}$.


We found that finger independence is coherent theoretical notions of articulatory effort such as handshape markedness: the 5 unmarked FS handshapes in \citet{battison1978lexical} (B, A, S, C, O) are among the 6 handshapes with the lowest finger independence in our data of 24 FS handshapes.




\subsubsection*{Handshape distance}

To quantify the perceptual effort of distinguishing between two handshapes, we measured the angular distance between each corresponding joint of the two handshapes. The more different the corresponding joint angles in two handshapes are, the more different the handshapes look from each other and thereby less perceptual effort is required by the observer to disambiguate between the two signs. 

This metric is similar to Keane's Articulatory Model of Handshape \cite{keane2014towards} where handshape similarity is measured by computing joint angle difference between canonical phonetic targets of handshapes, and has been shown to correlate with subjective similarity ratings by signers \cite{keane2017theory}. 

We define handshape distance as follows:
$$
HD(\text{hand}_1, \text{hand}_2) = \sum_{\alpha \in \text{hand}_1, \beta \in \text{hand}_2} D(\alpha, \beta) / N
$$

where for each joint, we computed the distance between the angle of the corresponding joint in the first and second handshape and took the mean over all joints.




\begin{figure*}[ht]
\captionsetup[subfigure]{justification=Centering}

\begin{subfigure}[t]{0.47\textwidth}
    \includegraphics[width=\linewidth]{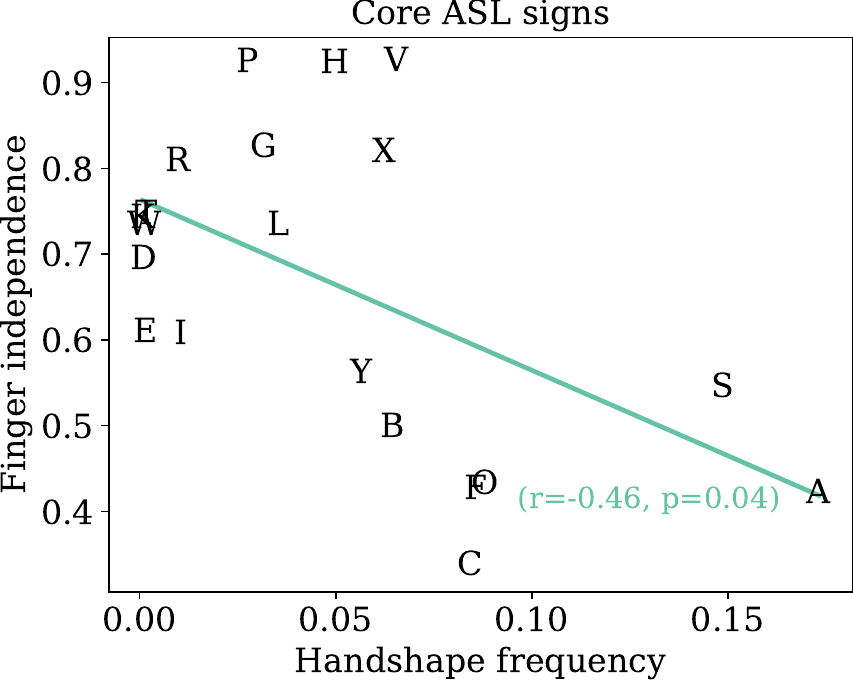}
    \caption{FS handshape frequency among native ASL signs vs. finger independence.}
    \label{fig:asl-freq}
\end{subfigure}\hspace{\fill} 
\begin{subfigure}[t]{0.47\textwidth}
    \includegraphics[width=\linewidth]{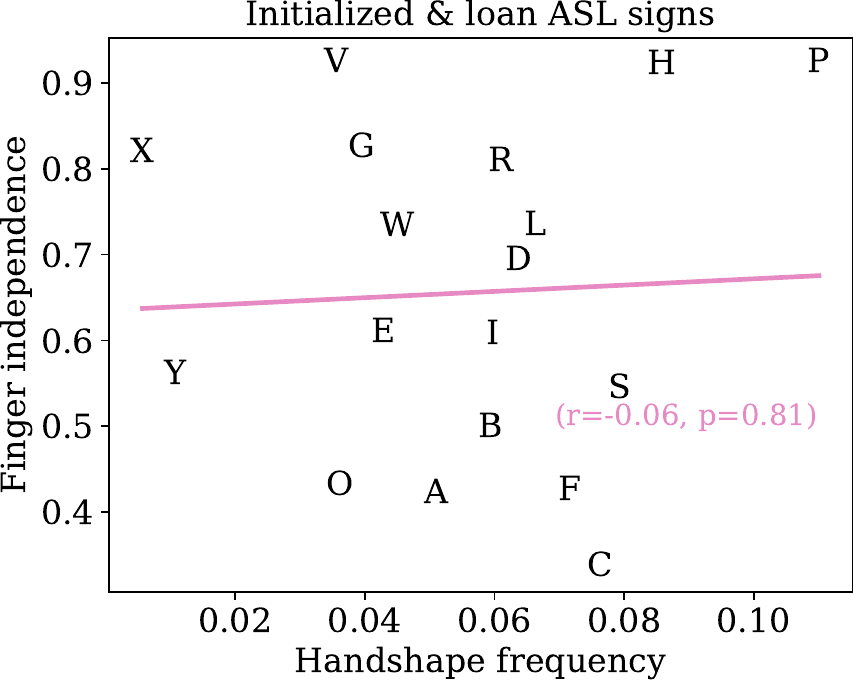}
    \caption{FS handshape frequency among initialized and loan ASL signs vs. finger independence.}
    \label{fig:for-asl-freq}
\end{subfigure}

\medskip 

\begin{subfigure}[t]{0.47\textwidth}
    \includegraphics[width=\textwidth]{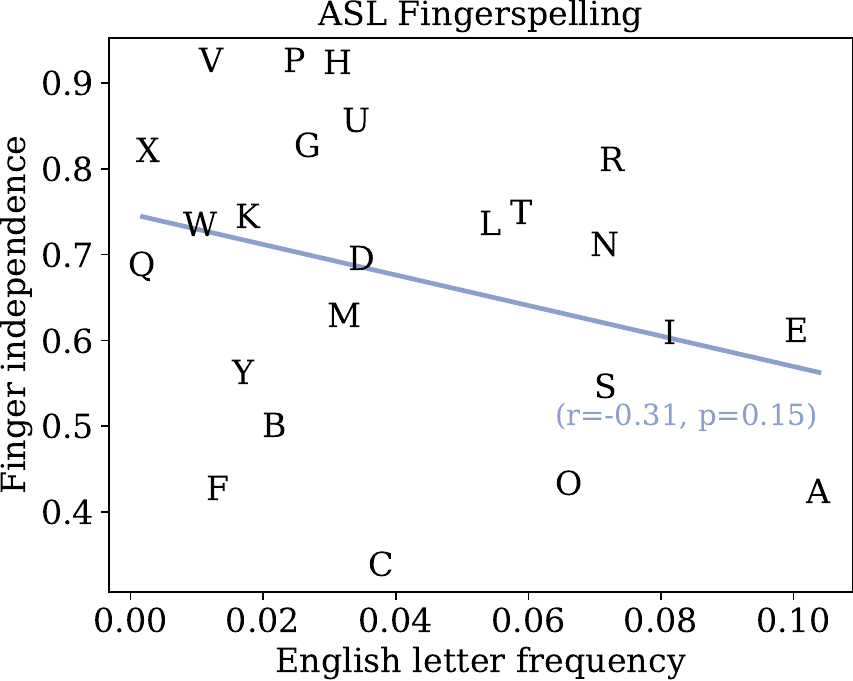}
    \caption{English letter frequency vs. finger independence.}
    \label{fig:en-freq}
\end{subfigure}\hspace{\fill} 
\begin{subfigure}[t]{0.47\textwidth}
    \includegraphics[width=\linewidth]{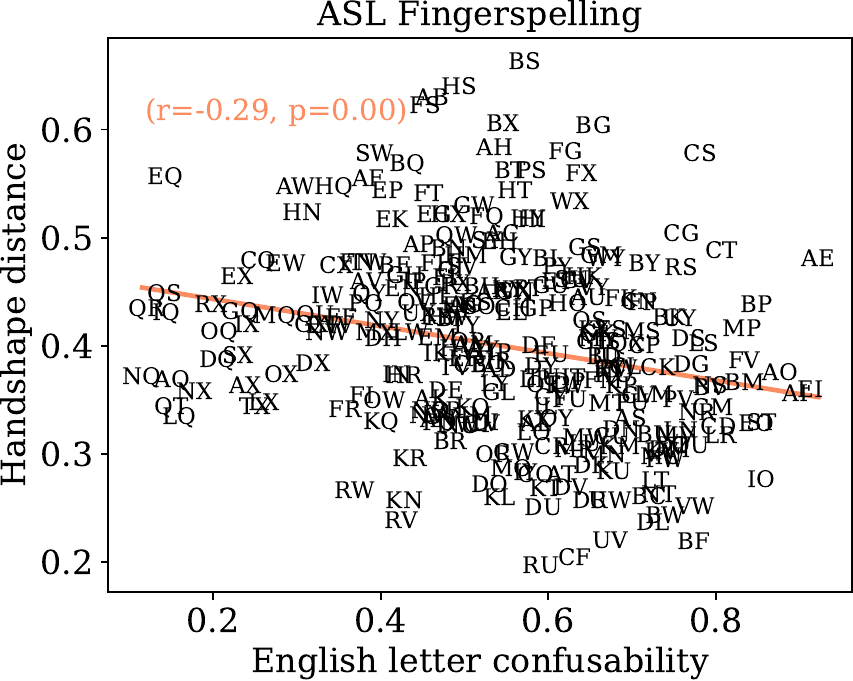}
    \caption{English letter confusability vs. FS handshape distance.}    \label{fig:conf}
\end{subfigure}
\caption{Correlations between usage statistics in ASL and English and articulatory/perceptual effort. FS handshape frequency in native ASL signs is inversely correlated with finger independence. However, there is no significant correlation between communicative effort and usage statistics from English.}
\label{fig:corr}
\end{figure*}

\section{Analysis}\label{sec:analysis}
To analyze the effects of articulatory and perceptual effort on ASL handshapes, we computed effort scores for each handshape in our dataset. Since our dataset contains several samples of each letter, and the number of samples is not constant across letters, for each letter we took the mean of the effort scores across samples. Then, we computed Pearson's correlation coefficient between effort scores and usage statistics using SciPy \cite{2020SciPy-NMeth}. We report and visualize correlations in Figure \ref{fig:corr}.

We summarize the three predictions we proposed in \S\ref{sec:intro}, following from our hypothesis that handshapes reflect pressure for communicative efficiency from ASL usage, but not from English usage:
\begin{enumerate}
    \item[\textbf{P1}]FS handshapes that appear frequently in native ASL signs are easier to produce.
    \item[\textbf{P2}]Letters that appear frequently in English are not necessarily easier to sign in ASL fingerspelling.
    \item[\textbf{P3}]Letters that appear in similar contexts in English are not necessarily easier to disambiguate from one another in ASL fingerspelling.
\end{enumerate}

\subsection{Articulatory effort}
We analyzed whether pressures for communicative efficiency are reflected in lower articulatory effort for high-frequency handshapes (\textbf{P1, P2}). First, we found a significant negative correlation between handshape frequency among ASL signs in the native component and finger independence ($r=-0.46, p=0.04$; Figure \ref{fig:asl-freq}), where frequent handshapes are easier to produce. This result supports \textbf{P1}.

On the other hand, we found no correlation between FS handshape frequency among ASL signs in the foreign component (initialized and loan signs) and finger independence ($r=-0.06, p=0.81$; Figure \ref{fig:for-asl-freq}).
There is also no significant correlation between English letter frequency and finger independence ($r=-0.31, p=0.15$; Figure \ref{fig:en-freq}), which supports \textbf{P2}. These results suggest that ASL reflects pressure for articulatory ease in FS handshapes, evidenced more clearly in native ASL signs than signs borrowed from English. 



\subsection{Perceptual effort}


We previously found no evidence for articulatory ease in signs borrowed from English. To further verify that English usage does not significantly impact communicative efficiency, we also examined perceptual effort in ASL fingerspelling (\textbf{P3}). 

We initially found a significant negative correlation between the confusability by the context of letters in English and handshape distance ($r=-0.29, p=0.00$; Figure \ref{fig:conf}). Although this result does not contradict \textbf{P3}, it is rather unintuitive: pairs of letters that appear in similar contexts are also more difficult to disambiguate in ASL fingerspelling, and there is a pressure for ``inefficiency''.

To better understand this unexpected outcome, we remark that many non-confusable pairs in Figure \ref{fig:conf} contain a low-frequency letter (Q, X, W), whereas pairs with high confusability are often composed of two high-frequency letters (A, E, N). We found that confusability by context and letter frequency\footnote{Here, we take the minimum of the frequency of each letter in the pair we measure.} are highly correlated ($r=0.69, p=0.00$): a pair of letters tend to be less confusable by context (i.e. tends to be able to be disambiguated based only on the context) when at least one letter in the pair is low-frequency. Intuitively, rare letters appear in a select set of contexts and are therefore less likely to overlap with the contexts of other letters.

We therefore measured the \textit{partial} correlation between confusability and handshape distance by partialing out letter frequency. We found that confusability and handshape distance are only weakly correlated ($r=-0.19, p=0.00$) once we control for letter frequency, suggesting that most of the contradiction is a result of the correlation between frequency and confusability. This also shows that there are no effects of perceptual optimization in FS handshapes from English confusability pressures, confirming \textbf{P3}.


\section{Discussion}


Revisiting our initial research questions, to answer \textbf{RQ1}, we found evidence for communicative efficiency in ASL handshapes, where \textbf{frequent handshapes are easier to produce} in native ASL signs. To answer \textbf{RQ2}, we did not find evidence of communicative efficiency in handshapes of foreign signs derived from English. Pressures for communicative efficiency are \textbf{mostly driven by ASL usage} rather than English usage.

There are a few possible explanations for this asymmetry between English and ASL pressures for efficiency. First, \citet{brentari2001language} demonstrated that foreign components of ASL obey fewer of the phonological constraints observed in the native component of ASL. Since fingerspelling and other foreign vocabulary lie at the periphery of the ASL lexicon, they may be less subject to the pressures for communicative efficiency that shape the native lexicon. 

Second, the ASL fingerspelling system traces its origin to the invention of hearing educators \cite{padden2003alphabet}, whereas native ASL signs evolve and diversify naturally through use by Deaf and signing communities \cite{woodward1973some, power2022historical}. This may lead to ASL signs being more likely to adopt linguistic forms that optimize communicative efficiency. 

Finally, ASL fingerspelling is relatively infrequent in signing discourse compared to native signs, constituting somewhere between 12 and 35\% of produced signs \cite{padden2003alphabet}. Studies on phonological change in spoken languages have shown that frequently used words undergo faster language change \cite{bybee2015language}. \citet{caselli2022perceptual} has also found that ASL signs observe similar effects, where more frequent signs reflect more optimization. Articulatory ease might play a stronger role in native ASL signs because of the latter's frequency of use.

\section{Related Work}\label{sec:related}

Relatively few works have examined theories of communicative efficiency in visual signed languages compared to studies on efficiency in spoken languages. \citet{ann2006frequency} studied handshapes in Taiwanese Sign Language (TSL) by manually categorizing TSL handshapes into three classes of articulatory ease, and found that more frequent handshapes are easier to articulate. We extended this analysis of the relation between articulatory effort and handshape frequency by designing an automatic way to quantify articulatory effort.

\citet{napoli2011some} studied efficiency pressures in ASL signing in casual conversations. They found that casual signing has a tendency for distalization, where it favors movement in arm and hand joints more distant from the torso than closer to the torso, presumably since moving a distal joint requires less energy than a joint close to the torso. We further investigated efficiency pressures on articulatory effort by studying handshape configurations rather than movements, and focusing on finger articulations in the dominant hand, rather than arm and hand joints.

\citet{caselli2022perceptual} compared the lexical frequency and handshape probability of 2000 lexical items in ASL to the location of articulation, and they found that signs with rare handshapes are signed closer to the face than signs with common handshapes, which optimizes perceptual ease since rare handshapes that are presumably harder to recognize are produced closer to where the receiver focuses their gaze. They also found that frequent signs are produced further from the face and closer to the resting position of arms than infrequent signs, which minimizes articulatory effort in producing frequent signs. We studied efficiency in handshapes more closely by quantifying articulatory effort in handshape configurations and perceptual effort in disambigurating between handshapes. In addition, we also explicitly compared how pressures for efficiency differ between native signs and foreign signs from ASL. 



\section{Conclusion}\label{sec:conclusion}

To conclude, our paper provides evidence of communicative efficiency in ASL handshapes: frequently used handshapes in native ASL signs require less articulatory effort. However, when examining ASL signs borrowed from English, we found no correlation between articulatory/perceptual effort in these handshapes and English usage. This suggests that the foreign component of ASL is not as influenced by efficiency pressures as the native component is, and optimizations for efficiency are mostly driven by ASL usage.

In the future, we aim to examine diachronic changes in ASL handshapes to determine if the handshapes of frequently used signs become easier to articulate over time. Additionally, we would like to investigate efficiency in signs borrowed from other signed languages, not just English, and to compare efficiency in the native and foreign components of signed languages other than ASL that also incorporate lexical borrowing. 



\section{Limitations}


The Google ASL Fingerspelling Recognition corpus released MediaPipe landmarks only to anonymize signers, so we did not have access to raw videos of ASL fingerspelling in our study. However, MediaPipe may contain inaccuracies in the position of landmarks, especially the z-coordinates where depth estimation is not always reliable, which may affect the accuracy of our effort metrics on ASL handshapes. To reduce inaccuracies from MediaPipe in our analysis, we sampled several examples of each handshape and take the average. We confirmed that our measured effort from this data is compatible with previous theoretical findings and our perceived effort. We also did not use palm orientation and wrist movement, so the lack of wrist rotational angle data in MediaPipe hand landmarks is not an issue. 

We also used infrequent English words from Wikipedia as an approximation of words fingerspelled in ASL, based on the assumption that words likely to be fingerspelled in ASL are proper nouns and specialized terms that are rare in English. 


\section*{Acknowledgements}
We thank Jenny Lu, Naomi Caselli, Alane Suhr, Nick Tomlin, Alex Shypula, Kaylo Littlejohn, Jiachen Lian, Gopala Anumanchipalli, and anonymous reviewers for helpful feedback. 
KY acknowledges support from the Future of Life PhD fellowship.

\bibliography{custom}

\appendix




\end{document}